\documentclass[manuscript,screen,nonacm,acmtog]{acmart}

\usepackage{xspace}

\setcopyright{acmcopyright}
\copyrightyear{2023}
\acmYear{2023}
\acmDOI{XXXXXXX.XXXXXXX}

\acmConference[KDD '23]{Knowledge Discovery and Data Mining}{August 6-10, 2023}{Long Beach, CA}

\newcommand{\eat}[1]{}
\newcommand{\hush}[1]{}

\newcommand{\Looper}{Looper\xspace}
\newcommand{\PEX}{PEX\xspace}
\newcommand{\Meta}{Meta\xspace}

\usepackage{microtype}
\usepackage{graphicx}
\usepackage{subfigure}
\usepackage{booktabs} 

\usepackage{hyperref}

\usepackage{amsmath}
\usepackage{mathtools}
\usepackage{amsthm}
\usepackage{bm}

\usepackage{multirow}

\usepackage[utf8]{inputenc} 
\usepackage[T1]{fontenc}    
\usepackage{hyperref}       
\usepackage{url}            
\usepackage{booktabs}       
\usepackage{amsfonts}       
\usepackage{nicefrac}       
\usepackage{microtype}      
\usepackage{bbm}
\usepackage[capitalize,noabbrev]{cleveref}

\theoremstyle{plain}

\theoremstyle{definition}

\theoremstyle{remark}

\begin{document}

\title{Scalable End-to-End ML Platforms:
from AutoML to Self-serve}

\author{Igor L. Markov}
\email{imarkov@meta.com}

\author{Pavlos A. Apostolopoulos}
\email{pavlosapost@meta.com}

\author{Mia Garrard}
\email{mgarrard@meta.com}

\author{Tianyu (Tanya) Qie}
\email{tanyaqie@meta.com}

\author{Yin Huang}
\email{maggiehuang@meta.com}

\author{Tanvi Gupta}
\email{tanvigupta@meta.com}

\author{Anika Li}
\email{lya@meta.com}

\author{Cesar Cardoso}
\email{cesarcardoso@meta.com}

\author{George Han}
\email{georgehan@meta.com}

\author{Ryan (Payman) Maghsoudian}
\email{ryanmag@meta.com}

\author{Norm Zhou}
\email{nzhou@meta.com}
\affiliation{
  \institution{Meta}
  \city{Menlo Park}
  \state{CA}
  \country{USA}
  \postcode{94025}
}

\begin{abstract}
ML platforms help  enable intelligent data-driven applications and maintain them with limited engineering effort. Upon sufficiently broad adoption, such platforms reach economies of scale that bring greater component reuse while improving efficiency of system development and maintenance. For an end-to-end ML platform with broad adoption, scaling relies on pervasive ML automation and system integration to reach the quality we term self-serve that we define with ten requirements and six optional capabilities. With this in mind, we identify long-term goals for platform development, discuss related tradeoffs and future work. Our reasoning is illustrated on two commercially-deployed end-to-end ML platforms that host hundreds of real-time use cases - one general-purpose and one specialized.
\end{abstract}



\maketitle
\pagestyle{plain}


\section{Introduction}

Many core techniques for artificial intelligence (AI), especially machine learning (ML), are available today and can be invoked easily via open-source software \cite{gulli2017deep, pedregosa2011scikit} or cloud services \cite{elger2020ai, salvaris2018microsoft}. However, connecting them to software products, training on appropriate data, and driving business value remains challenging.

The McKinsey State of AI 2022 report \cite{mckinsey2022} notes \cite{vb2022} that industry AI/ML adoption more than doubled since 2017 and then leveled off, with common concerns on how to scale up the use of AI and speed up AI development. The report breaks down AI/ML adoption by capabilities and use cases, the latter being led by Service Operations Optimization (24\%), Creation of New AI-based Products (20\%), Customer Service Analytics (19\%), Customer Segmentation (19\%), New AI-based Enhancement of Products (18\%), etc. The sum over all categories is ~180\%.
A global survey on AI adoption by 2025 — available in the “CIO vision 2025: Bridging the gap between BI and AI” report by MIT Technology Review \cite{ciovision2025} — reveals several important trends. Out of 600 CIOs and other technology leaders polled, AI is used today by almost all companies represented (94\%+ in each of seven enterprise categories). Of those, an overwhelming majority (84\%) of companies are not AI-driven and remain in early stages of AI adoption. Perhaps, such statistics can be extrapolated to individual large companies as well, where a handful of AI-driven applications are surrounded by numerous AI-light applications. When asked to identify  the top priority of their enterprise data strategy for the next three years, 78\% of the respondents chose scaling Al and machine learning use cases to create business value. In particular, a wider adoption of AI to a variety of use cases is seen as mission-critical.

\noindent
{\bf Data-driven software platforms}. The McKinsey State of AI 2022 report notes that “most respondents indicate  that hiring for each Al-related role has been difficult in the past year and hasn't become easier over time.” Rather than hire such engineers and scientists for each application, we advocate building data-driven software platforms that orchestrate components, services, and access interfaces to support a variety of external or platform-hosted products by implementing and automating workflows to perform application-specific tasks. In other words, a more efficient use of precious expertise is to increase the accessibility of applied AI by developing software platforms that lower the barrier to entry.
Another benefit of data-driven software platforms is illustrated by a common strategy \cite{molinaro2022privacy} is to limit retention of data and trained ML models to, say, 35 days, which improves data privacy but requires infrastructure for feature lineage tracking and automation for ML model retraining.

\noindent
{\bf ML platforms} \cite{Hermann2017Michelangelo, markov2022looper, soifer2019inference} often support and automate workflows that train ML models on data to perform prediction, estimation, ranking, selection, and other ML tasks. Such applications necessitate regular data collection and retraining of ML models \cite{wu2020deltagrad}, which provides strong motivation for platforms in practice. More frequent retraining is common when data trends change quickly. Trained models must be hosted for batch-mode or real-time inference. Different applications call for different ML model types, and a broad portfolio of auto-configurable components may bring strong results faster than hand-optimizing a small set of components. Choosing between various model types (tree-based, DNN, etc) and combining them encourages automation.

 ML platform development is driven by the following tradeoffs:
\begin{enumerate}
\item Customer needs vs. technology availability,
\item Expert support, fine configuration, and optimization for high-end applications vs. automation to reduce cognitive load and engineering effort,
\item System-level development (driven by traditional SW engineering, design and architecture considerations) vs. ML-driven development (initiated and driven by data, model and metric considerations),
\item Reactive short-term efforts vs. proactive long-term plans,
\item Orchestrating numerous point tools into entire workflows.
\end{enumerate}

\noindent
{\bf End-to-end ML platforms} support workflows with a broader scope, including data collection and preparation as well as tracking and optimization of product-impact metrics to drive business value. While relatively new, several industry platforms \cite{markov2022looper, Hermann2017Michelangelo} drive products that are used by billions of users and operate on powerful data sources. Such platforms automate data collection and model retraining, and also support online causal evaluation for products they enable. A/B testing \cite{kohavi2017online, gilotte2018offline} with product metrics is a particularly big step towards evaluating end-decisions driven by ML models instead of only evaluating ML models in terms of closed-form loss function(s) on pre-saved data \cite{gomez2015netflix}, as is common in traditional ML platforms. End-decisions affect product metrics and usually cannot be handled by closed-form functions, and evaluating product impact of multiple alternative changes requires causal product experiments (otherwise, there is no guarantee that correlations captured by ML models have a significant and positive impact on product metrics). A/B testing captures a variety of phenomena \cite{fabijan2017benefits} that are difficult to model and predict, such as network effects.

End-to-end platforms are categorized as specialized and general-purpose \cite{markov2022looper}. Specialized platforms focus on certain application categories and use cases (e.g., image \cite{khan2021machine}, text \cite{chowdhary2020natural}, or speech analysis for harmful content \cite{nassif2019speech}), which allows them to tailor their interfaces to application concepts and product specifics. They may support a handful of high-value customers or a larger number of customers interested in such particular ML capabilities. General-purpose platforms offer broader capabilities and the flexibility of configuration, but often must justify the engineering investment via sufficiently broad adoption during platform development.

\noindent
{\bf Scaling the adoption of end-to-end ML platforms and driving the business value in applications} are the main challenges discussed in this work. Thus, we upgrade end-to-end platforms to {\em self-serve platforms} and describe this new concept in detail in Section \ref{sec:self-serve}. Achieving the {\em self-serve} quality requires pervasive use of AutoML techniques (Section \ref{sec:background}), platform integration, online testing, as well as balancing the five tradeoffs for ML platforms itemized above. We illustrate this approach for two high-volume commercial end-to-end ML platforms and outline future work necessary for an effective upgrade (Sections \ref{sec:details} and \ref{sec:conclusions}). Deployed for several years, today these platforms support millions of AI outputs per second: \Looper is a general-purpose platform, while \PEX is relatively more specialized. Recent platform improvements toward self-serve are discussed in Section \ref{sec:deployment} along with deployment experience.

\section{The landscape of ML automation}
\label{sec:background}
Most of the literature on applied ML focuses on what we call the Kaggle paradigm, where training data and loss functions are provided, and the goal is to train a model that, when evaluated on hold-out data, would minimize the loss function. To this end, the Kaggle Web site hosts competitions \cite{yang2018deep,puurula2014kaggle} for creating such ML models and determines the winners by loss function values. Such competitions help evaluate new model architectures and, less often, new optimization algorithms, but don’t capture implementation and runtime tradeoffs, where a slight technical win may come at the cost of hand-tuning and customized optimization, not to mention model size, inference latency, etc. To bridge the gap between Kaggle competitions, a newer field of MLOps \cite{makinen2021needs} combines ML with DevOps (development operations, \cite{leite2019survey}) and focuses on deployment and maintenance of optimized ML models in applications. A key issue for industry ML deployments is the efficiency and overall capacity of inference \cite{soifer2019deep}. However, most of the published MLOps work \cite{kreuzberger2022machine} is limited to model-based metrics, such as the loss function(s), input and output distributions, etc. We find it convenient to view the Kaggle paradigm broadly as including such MLOps considerations.

\noindent
{\bf AutoML toolboxes}. Within the Kaggle paradigm, automation is needed to optimize hyperparameters \cite{yu2020hyper} for ML models and perform neural architecture search \cite{elsken2019neural} for deep learning models. These tasks can be accomplished using traditional optimization algorithms, novel applications of deep learning to optimization, and specialized methods to optimize deep-learning models (such as training supernets). For practical uses, AutoML techniques \cite{he2021automl} must be implemented using a consistent interface that supports interchangeability, composition, ML pipeline management, and live-product experimentation.  AutoML toolboxes that offer such implementations can be illustrated by open-source libraries, such as:

\begin{itemize}
\item \href{https://github.com/microsoft/FLAML}{FLAML} (Microsoft, \cite{Wang_FLAML_A_Fast_2021}) - a lightweight AutoML library that handles common tasks (model selection, neural architecture search, hyperparameter optimization, model compression),
\item
\href{https://github.com/google/vizier}{Vizier} (Google, \cite{oss_vizier}) - a uniform Python interface for black-box optimization intended for hyperparameter optimization that offers a variety of optimization algorithms,
\item \href{https://ax.dev/docs/why-ax.html}{Ax} (Meta, \cite{botorch}) – a general ML tool for black-box optimization that allows users to explore large search spaces in a sample-efficient manner for services such as multi-objective neural architecture search, hyperparameter optimization, etc.
\item \href{https://automl.github.io/auto-sklearn/master/}{Auto-sklearn}\cite{feurer-neurips15a, feurer-arxiv20a} - an automated machine learning toolkit and a drop-in replacement for a scikit-learn estimator,
\item \href{https://github.com/autogluon/autogluon}{AutoGluon} (Amazon, \cite{agtabular}) - an AutoML library that manages a variety of ML models, with provisions for image, text and tabular data, as well as multimodal data,
\item
\href{https://github.com/IBM/lale}{LaLe} (IBM, \cite{baudart_et_al_2021}) - an AutoML library for semi-automated data science,
\end{itemize}

\noindent
{\bf A high-level snapshot of the field} can be seen in
\cite{hutter2019automated}
 and, more broadly, in the \href{https://automl.cc/call-for-papers/}{Call for Papers} for the 1st International Conference on AutoML (AutoML-Conf 2022) that lists tracks for submitting papers from industry and academia. We give a condensed version:

\begin{itemize}
\item Neural Architecture Search (NAS, \cite{elsken2019neural}) and Hyperparameter Optimization (HPO, \cite{yu2020hyper})
\item Combined Algorithm Selection and Hyperparameter Optimization (CASH)
\item AutoML algorithms \cite{he2021automl}: Bayesian Optimization, Multi-Objective
\item AutoAI (incl. Algorithm Configuration and Selection, \cite{cao2022autoai}) and Meta-Learning \cite{vanschoren2018meta}
\item Automated Data Mining
\item Reproducibility
\item Automated Reinforcement Learning (AutoRL, \cite{parker2022automated})
\item Trustworthy AutoML with respect to fairness, robustness, uncertainty quantification and interpretability \cite{amirian2021two}.
\end{itemize}

\noindent
{\bf AutoML capabilities in end-to-end ML platforms} are facilitated by AutoML tool-boxes (frameworks). However, in addition to the Kaggle paradigm,  these capabilities also help manage the ML lifecycle from data collection to product impact tracking and are invoked as services. When integrated with an ML platform and usable with no ML coding, they form platform-level AutoML solutions. Priorities for AutoML services can be application-specific, and also emphasize integration in addition to component quality. Before we discuss specific end-to-end systems, we note that \cite{breck2017ml} illustrates the significance of automation in production systems. This paper presents “28 specific tests and monitoring needs, drawn from experience with a wide range of production ML systems to help quantify these issues and present an easy to follow road-map to improve production readiness and pay down ML technical debt.” Those tests require proper formalization of interfaces, explicit capture of metrics and requirements, reproducibility and automation of tests, full hyperparameter tuning, full automation of data pipelines, automatic canarying of candidate ML models before promotion or rollback, automatic model refresh and promotion, etc. This is why our overall strategy in this work focuses on pervasive use of AutoML techniques, platform integration, and online testing.

\section{Two End-to-end ML platforms} 
\label{sec:twoplatforms}

We now briefly review two end-to-end ML platforms on which we later illustrate our proposed concept of a self-serve platform. Almost 100 product teams across currently leverage them to drive product-metric improvements. These platforms are particularly relevant because several of their architectural aspects bring them halfway through to this concept.
\begin{itemize}
\item
\Looper and \PEX maintain full custody of data — from (i) automatic data collection to (ii) causal model evaluation by means of A/B testing with product metrics (no “clean” data is required). The context and APIs for automated data collection differ between the two platforms, but the results are similar: simplified maintenance, reduced engineering effort by eliminating known pitfalls with data collection.
\item
\Looper and \PEX do not require customers to define or implement new model types. A variety of managed model types can be chosen from or auto-selected, and are then interfaced with the entire ML workflow without the need for customer-written ML code.
\item
\Looper and \PEX platforms are config-driven, maintain reproducible ML models and regularly retrain ML on fresh data to adapt to data drift. Retrained models are evaluated and promoted (automatically) upon positive results.
\end{itemize}

Each of the two platforms enjoys a distinct broad customer base among product teams and is proven to drive business value in numerous user- and system-facing applications. Together they produce 3-4M AI outputs per second for hundreds of product use cases. A detailed architectural description of the platforms and a discussion of applications are available in [removed for blind review], but in this section, we highlight key technical aspects, important use cases and how product impact is measured.

\Looper and \PEX specialize in tabular data, where features may be of different types and scales, might not correlate (like image pixels do), and might include individually-engineered features (like event counters). AI outputs of \Looper and \PEX often choose between several options, e.g., in binary decisions or when determining configuration options for various software products. In \Looper, binary classification models are also trained for other applications, such as ranking, where scores produced by those models are sorted to order given items.

\noindent
{\bf \Looper} is a general-purpose end-to-end ML platform that enables  product teams to deploy fully managed real-time ML-based strategies to optimize impact metrics for a variety of product surfaces. It obtains training data through a predict-observe API that observes desired results shortly after the predictions and joins them for use when retraining models. \Looper back-end supports a variety of ML tasks (including classification and regression, supervised and reinforcement learning, multimodal-multitasks and contextual bandits) and model types, but does not require that product engineers have prior ML experience or even understand the different model types. By lowering barriers to entry, \Looper broadens the use of ML in \Meta products and simplifies maintenance for ML-based applications.
The typical time spent to launch a model is reduced to one month from several months for traditional ML development cycles.

\Looper performs multiple stages of optimization within the platform. In the first (offline) stage, loss functions are optimized during model training, and hyperparameters are optimized using offline Bayesian optimization. When a resulting model is deployed, the platform tracks and optimizes the product metrics. In the second (online) stage, product metrics can be improved further by methods unrelated to traditional loss functions, e.g., by tuning decision policies that postprocess the outputs of ML models.

\noindent
{\bf \PEX} 
addresses ML needs for in-situ live-product experimentation within multiple product surfaces. It is a specialized end-to-end ML-backed experimentation framework that enables product teams to leverage \href{https://egap.org/resource/10-things-to-know-about-heterogeneous-treatment-effects/}{Heterogeneous Treatment Effects} (HTE, \cite{kunzel2019metalearners}) to optimize end-user experience at the level of individual end-users. \PEX predicts (based on user features) which variant treatment performs  best for a given end-user by obtaining training data directly from the results of A/B tests (viewed as training labels) and can launch sequences of A/B tests to optimize hyperparameters.  Traditional live-product experimentation relies on the Average Treatment Effect (ATE) to determine a single static decision that performs best on average to all users. But ATE often does not perform best for all users, meaning some users are receiving a worse experience than others. In contrast, \PEX learns to predict product metrics for all the possible treatment assignments for a given end-user. Depending on the context and needs, \PEX trains

\begin{itemize}
\item
An HTE meta-learner that more accurately models the difference between immediate effects on the product metrics (techniques for this ML task are given in \href{https://arxiv.org/abs/1706.03461}{arXiv:1706.03461}).
\item An RL model, adept at optimizing over a longer time horizon and capable of modeling delayed product metrics for multiple alternative treatments, as well as of choosing the most promising treatment. RL models are generally more suitable for use cases where end-users interact repeatedly with the product.
\end{itemize}

\PEX then builds decision policies to provide each user with a personalized treatment. A decision policy either post-processes the outputs of an ML model (as in the case of HTE meta-learners) or is directly expressed by an ML model (as in the case of RL).

\PEX performs policy optimization in two phases: offline and online. First, \PEX ensures  the given use case is eligible for personalization, by running offline policy optimization with counterfactual evaluation that simulates responses that would be observed online. Assuming eligibility, the offline policy optimization (driven by black-box optimization methods) identifies the “best” (for a single metric) or several non-dominated (for multiple metrics) policies to be evaluated online. Such early pruning helps avoid online evaluation (A/B tests) of poor-performing policies (and thus avoid product-experience deterioration for some end-users). In the second phase, the candidate policies are evaluated online and each is optimized further using sequential live-product experimentation framework Adaptive Experimentation \cite{bakshy2018ae} which uses Bayesian Optimization to iteratively improve policy configurations.

\PEX is tightly integrated with a company live-product experimentation system, while supporting sufficient abstraction level to provide the most intuitive user experience. Prior ML knowledge is not necessary to successfully run \PEX and launch new decision policies in production. Since 2020, there have been 40+ product launches from teams across \Meta. \PEX helps \Meta engineers to maximize product impact by personalizing experiences via ML, with moderate live-product experimentation effort.

\noindent
{\bf The value proposition} of end-to-end ML platforms like \Looper and \PEX is that shared engineering effort (new technologies, regular platform maintenance, and system upgrades) can help customers focus on applications. AutoML plays a significant role in supporting this value proposition by scaling configuration and optimization. Both platforms are integrated with company infrastructure and especially other related platforms.

\section{Self-serve ML platforms and tradeoffs}
\label{sec:self-serve}

End-to-end ML platforms have shown their worth by supporting numerous product use cases, improving product metrics and/or reducing maintenance costs. A brief review of industry end-to-end platforms can be found in [removed for blind review]. Further development of such platforms is driven by attempts to scale them to more applications and support hundreds of platform customers with minimal engineering support to individual customers. However, minimizing support to individual customers can only be achieved via engineering investments into platform development. This paradox is at the core of our exploration, and we develop guidelines for such investments. Specifically, we introduce the notion of a self-serve ML platform, supported by an integrated suite of AutoML techniques (prior work on AutoML focuses on specific subsets of the ML lifecycle). We then discuss this notion for the \Looper and \PEX platforms specifically to illustrate how it may be applied to other platforms.

\subsection{The notion of a self-serve platform}
\label{sec:notion}

We studied the entire lifecycle of ML development in end-to-end platforms and identified steps that must be automated for the platforms to qualify for self-serve. Some of the greatest challenges and, respectively, the greatest opportunities are associated with data handling and with product impact.

\subsubsection{Data handling}
Maintaining a full custody of data (from API-driven data ingestion and collection of training data to inference and in-situ product evaluation) and automating related operations saves effort and avoid many human errors, inconsistencies and pitfalls.

\subsubsection{Product impact evaluation and optimization}
Given data sources and preprocessing, as well as ML models and decision policies, we are interested in
\begin{itemize}
\item {\em Observational tasks}: evaluating, learning, and modeling the impact of AI outputs on product end-metrics,
\item {\em Interventional tasks}: optimizing prediction mechanisms (models, decision policies) to improve product metrics.
\end{itemize}

\noindent
{\bf Observational tasks} for product impact are less direct (compared to loss-function evaluation when training individual ML models) and require additional considerations.
Longer-term product metrics implicitly aggregate the impact of AI outputs over time;
Product impact is causal in nature and depends on external factors (users, systems, environments) that often cannot be fully modeled; non-causal methods (such as traditional ML models based on correlations) are unable to estimate some causal effects;
When estimating causal effects with external factors, any one of multiple possible AI outputs can be evaluated but typically no more than that — a user would not walk through the same door twice, and a system will not perfectly replay a sequence of events to try different responses.
Live-product causal experiments (such as A/B tests) consume valuable resources and must use available data effectively. Otherwise, they risk being underpowered and fail to produce statistically significant results. Before live-product experiments, clearly-deficient candidate ML configurations are caught by offline evaluation using proxy metrics.

\noindent
{\bf Interventional tasks} require more effort relative to offline evaluation, experiment cycles may be long to have enough data to make a judgment, and there is always the risk of deploying terrible experiments. The direct approach to circumvent these is counterfactual evaluation, as it allows estimating the outcomes of potential experiments without actual deployment, and can drive offline optimization based on the effect of imaginari stimuli for identifying good candidates for online evaluation.

Observational and interventional tasks require time and expertise, and carry risks. Hence, the management of live-product experiments and product impact optimization must be automated.

\subsubsection{Requirements for a self-serve platform}
Informally, a self-serve platform enables customers to deploy effective applications with little support. We expand on this intuition and list various aspects of platform development necessary to back it up. Appropriate metrics may be developed for individual platforms and contexts.

 \noindent
{\bf Self-serve ML platforms} must satisfy the following requirements:
\begin{enumerate}
\item
{\bf low cognitive barrier} to entry and low requirements for ML experience for product engineers,  (in addition to “hiding” routine tasks behind automation; the UI should avoid unnecessary dependencies on ML concepts, data science concepts, as well as \Looper and \PEX architecture concepts; when unavoidable, these concepts can be explained in tooltips and/or using links to documentation).
\item
{\bf automated data collection} from applications and customization of subsequent data preprocessing (normalization, outlier removal, data imputation, down/up sampling, etc)
\item
{\bf AutoConf}: automated selection of an ML problem formulation, ML task (ranking, classification, etc), model type and default parameters, followed by workflow-automated traditional AutoML (parameter selection, network architecture search, etc),
\item
{\bf product-impact metrics} - tracking and automatic optimization of product-impact metrics, support for (i) counterfactual evaluation and (ii) online causal evaluation, such as A/B testing
\item
{\bf sufficient ML quality} with limited manual configuration and optimization effort
(comparisons are made to (a) custom AI solutions, (b) AutoML tools and services)
\item
{\bf full management of hosting} of data, models and other components, with modest resource utilization (some customer resources may not be accessible),
\item
{\bf adaptation to data drift (calibration and retraining)} to ensure model freshness,
\item
{\bf resilience and robustness to disruptions in data (delayed, missing) and system environment (resource limitations and
outages)} with minimal recurring customer-side maintenance effort,
\item {\bf customer-facing monitoring} and root-causing of customer errors
\item
{\bf scalable internal platform maintenance} and customer white-glove support.
\end{enumerate}

Attaining self-serve quality includes point-automation as in traditional AutoML, flow-automation which connects point optimizations, as well as whole-strategy (vertical) optimizations that co-optimize multiple point components. It would be naive to support self-serve by a custom-engineering service backend — instead support for individual customers is either generalized or traded off for more scalable platform support (provided by platform development engineers). An apparent limitation is that the largest ML applications may require too much custom optimization and maintenance. On the other hand, a self-serve platform helps even seasoned ML engineers by automating boring and error-prone routines for less critical but numerous ML applications.

\noindent
{\bf Additional capabilities} for self-serve platforms depend on the client base and uses:
\begin{enumerate}
\item
open architecture: in addition to the end-to-end ML support, the self-serve platform may offer its components and partial workflows individually (such as model-training for client-provided data or using a client-provided model), such decomposability helps leverage ML platforms with infrastructure client-designed for high-value use cases.
\item
customizations to common ML tasks: ranking and selection — succinct high-level APIs that use relevant concepts, support for relevant model architectures, loss functions, regularizations, output constraints, and diagnostics, etc.
\item
reproducibility of models (i.e., all necessary code and data are available; where actual training data may be subject to retention policies, comparable data are available).
\item
meta-learning, including transfer learning - automatically choosing learning parameters, reusing and adapting trained models to new circumstances.
\item
interpretable ML models -- ability to provide platform client insight into model behavior, without requiring the understanding of model internals.
\item
fairness in ML addresses various ways to evaluate fairness (proportionality, equal reach, opportunity, impact, etc) and ways to train ML models to improve these metrics.
\end{enumerate}

These capabilities broaden adoption in several ways. Open architecture facilitates reuse, evaluation and improvement of individual platform components. Clients with ML experience would benefit from open architecture more than entry-level clients that may prefer end-to-end support. Customizations to common ML tasks help platform clients to support certain applications. Reproducibility of models maintains a healthy SW and ML development cycle, helps keep models fresh, and also supports recurrent retraining of ML models. Meta-learning, including transfer learning, seeks more effective models and faster training, based on past experiences. Explicable ML models play an important role in personalized treatment to avoid inadvertent biases, receive legal approval, and provide platform users greater understanding of their client base. Fairness support is required in certain applications, but remains challenging when no clear guidelines are provided given the various notions of fairness that exist.

\begin{table*}
\caption{
\label{tab:selfserve}
Applying the notion of self-serve to the \Looper and \PEX platforms.}
\begin{tabular}{|c|c|c|}
\hline
\sc 10 requirements & \sc \Looper & \sc \PEX \\
\hline
Low cognitive barrier to entry &
Broader scope of UI &
Specialized UI and data sources \\
\hline
Automated data collection &
Generic \Looper APIs &
\parbox{5.5cm}{
Labels from A/B experiments; \\
API to experimentation system} \\
\hline
AutoConf &
\parbox{5.5cm}{
Selection of ML task, model type, features, default params;
Hyperparameter tuning
Decision-policy tuning for binary classification;
Value-model tuning for multiclass classification and multimodel multitasks;
}
&
\parbox{5.5cm}{
Fixed ML task, selection of HTE meta-learner, RL, or derived heuristic policy, feature selection, offline/online policy optimization and param tuning }
\\
\hline
Product-impact metrics &
\multicolumn{2}{c|}{
Tracking and optimization are critical in most applications} \\
\hline
Sufficient ML quality &
\parbox{5.5cm}{
Last 1-2\% model quality viz. loss functions (SOTA) is not critical in many cases. Overall quality is most affected by the decision policies, and then by the ML model.} &
\parbox{5.5cm}{
Somewhat more important than for \Looper. Greatly affected by label selection} \\
\hline
Full management of hosting &
\multicolumn{2}{l|}{
\parbox{11cm}{
Clients use their storage quota for data, pipelines, but management is fully automated. This includes automatic canarying and promotion of retrained models.
}} \\
\hline
Adaptation to data drift &
\multicolumn{2}{l|}{
\parbox{11cm}{
For nonstationary data: automatic model calibration, model retraining and promotion; decision policy re-tuning.
}} \\
\hline
Resilience, robustness to disruptions &
\multicolumn{2}{l|}{
\parbox{11cm}{
Data distribution monitoring, real-time alerts;
handling of missing data
}} \\
\hline
\parbox{4.5cm}{
Client-facing monitoring and\\ root-causing of client errors.}
&
\multicolumn{2}{l|}{
\parbox{11cm}{
Alerts on anomalies in data and AI outputs.
Help diagnose and address client and platform errors.
}}
\\
\hline
Scalable internal platform maintenance &
\multicolumn{2}{l|}{
\parbox{11cm}{
Maintenance load is due to: (i) company-wide system environment changes, (ii) client activity, (iii) use-case issues: quotas, missing data, etc.
}} \\
\hline
\multicolumn{3}{c}{}
 \\
\hline
\sc 6 additional capabilities & \sc \Looper & \sc \PEX \\
\hline
Open architecture &
\parbox{5.5cm}{
 Training models on given data/comparable to Google AutoML tables;
Logging as a separate service}
&
N/A
\\
\hline
Customizations to ML tasks &
\parbox{5.5cm}{
Ranking as a service, selection (top N out of a large predefined set), etc}
&
\parbox{5.5cm}{
The formal ML task is fixed, but outputs are used for a variety of applications (UI or value model optimizations, etc).
}\\
\hline
Reproducibility of models &
\multicolumn{2}{l|}{
\parbox{11cm}{
Automatic model refresh (i) to adapt to data drifts, \\
(ii) to meet data retention limits related to data laws.
}} \\
\hline
Meta- and transfer learning &
\parbox{5.5cm}{
Transfer-learning across related apps, where some training data can be reused}
&
\parbox{5.5cm}{
Meta-learners (T learner, X learner, etc)
}
\\
\hline
Interpretable ML &
\parbox{5.5cm}{
Feature importance analysis (mostly for model building);
Monotone constraint for model output w.r.t. feature values.
}
&
\parbox{5.5cm}{
In addition to feature importance analysis, automatic user segmentation analysis provided to understand meta-learners;
Automated heuristic policy distillation possible
}
\\
\hline
Fairness in ML &
\multicolumn{2}{l|}{
Pending guidance for individual applications
}
\\
\hline
\end{tabular}
\end{table*}

In terms of the overall impact of self-serve platforms, we notice several tradeoffs. In particular, scaling client adoption and lowering the barrier to adoption increases platform-level maintenance load. Another source of additional work is interoperability with other platforms and company-wide systems (revision control and build systems, basic ML environments, experimentation platforms, etc) - serious upgrades or changes of interface in partner systems require attention and urgent efforts. Thus, scaling internal maintenance becomes important. Achieving self-serve quality and scaling adoption also requires additional platform development, advocacy to attract product teams, better tracking of client roadmaps and impact metrics, and some amount of white-glove support for high-impact clients. There are important tradeoffs among model configurability, model optimization, and eligibility for our platform, as well as operational decisions about the amount of engineering support offered to individual clients, etc. In particular, most valuable application-specific models are supported by a large cadre of engineers that try to optimize and extend those models in parallel, and such work necessitates flexible configuration infrastructure that can be difficult to support in a general platform. When a model or a use case requires  white-glove support from the platform team, the support effort may be better spent on platform improvements for multiple clients. In such cases, we try to generalize lessons from one use case.

While reaching self-serve quality is a worthy goal, its value is greatly increased upon attaining an economy of scale. For the latter, we give the following criterion: very few client-support and maintenance activities are done for individual clients.

\section{Extending ML Platforms toward Self-serve}
\label{sec:details}

The ten stated requirements for self-serve platforms and six additional capabilities apply to \Looper and \PEX in somewhat different ways, given that \Looper is a general-purpose end-to-end platform while \PEX is a more specialized platform.
Table \ref{tab:selfserve} contrasts relevant differences.
For example, automated data collection in \Looper is implemented via the \Looper APIs, which are used by many diverse applications and from several programming language environments. For \PEX, despite a variety of specific product applications, training data for ML models is collected from A/B experiments via integration with \Meta’s widely adopted online experimentation system. To ensure sufficient ML quality, \Looper offers several types of ML models, including tree ensembles, deep learning, contextual bandits, reinforcement learning, etc, along with extensive feature selection.
 The \PEX platform is directly integrated into
the company-standard experimentation system to
reduce the cognitive barrier to entry. Additionally,
we leverage consistent UIs and APIs to provide a seamless transition from randomized experimentation to personalized experimentation. We also surface our client-facing monitoring and alerting via integrated UI.

\section{Improvements and deployment experience}
\label{sec:deployment}

Three implementation efforts brought the \Looper and \PEX platforms closer to {\em full self-serve}. To this end, we also describe deployment experience, impact on the usability of the platforms, and client survey results. Improving ML performance is not a primary objective of these efforts, but small ML performance improvements were observed in some cases as well.

\subsection{Product-metric optimization in \Looper \& \PEX}
\label{sec:pismo}

For applications deployed on our platforms, models are auto-retrained when new training data is available, but decision policies were not. A handful of clients retune decision policies manually, but most teams don’t due to limited engineering resources and the tedious effort required.
A survey of platform clients indicated high importance of unlocking regular improvement of product metrics by updating smart strategies (including ML models and decision policies) as data trends change. This can
\begin{enumerate}
\item Save significant manual effort to run experiments when optimizing smart strategies
\item Enable automatic configuration and launching of online experiments to relianbly tune decision policies
\item Increase trust with clients by accurately tracking and optimizing product metrics via the platform
\end{enumerate}

To realize these possibilities, we now allow platform clients to specify the product decision space upfront: the metrics to optimize, the experimentation preference for each product decision (such as whether to prefetch, whether to send notification), the relations among product decisions, etc. This information is then used to automatically generate decision policies or launch offline analyses and efficient online experimentations to tune the smart strategies. After smart strategies are tuned/set up, \Looper and \PEX can generate product decisions directly for the incoming traffic.
After enabling automatic decision policy tuning,
we’ve observed the following benefits:
\begin{itemize}
\item Unlock {\em repeated} product metrics improvement. We validated this workflow on several examples. We onboarded a use case which determines whether to prefetch certain content (such as stories, posts, reels) to a given edge device (smartphone) during peak hours. We launched newly-tuned decision policies which improved the team’s top-line product metrics for different device types. The overall success rate is improved by 0.73\% with neutral comptational cost.
\item
Save at least two weeks of manual effort for running experiments to maintain decision policies each half for each loop/use case. So far, we saved at least six weeks of total manual effort for three onboarded loops.
\item
Automate and enable product metrics evaluation on decision policy iterations, otherwise product-metric impact of decision policy improvements cannot be justified systematically.
\end{itemize}
Automatic model tuning and refresh have been deployed for several months without disruptions.
These efforts brought platform automation to a new level ---
after training ML models, the clients no longer need to build and maintain the decision policies on their own. They can rely on the platform to automatically improve product decisions over time
with all necessary safeguards.

\subsection{Client-driven platform improvement for \PEX}
\label{sec:selfservepex}

To evaluate the self-servability of the \PEX platform, we conducted a User Research Study where 6 users from 6 different teams were interviewed about their experience using the platform. All users indicated a strong favorable sentiment of the value of \PEX with respect to to product-metric improvement, and a strong interest in self-servability for ML platforms in general. At least some level of unfamiliarity with ML was ubiquitous across all users, and some findings pointed to the need for even further reduction of the cognitive barrier to entry. For example, feature selection was a major pain point for most clients interviewed as they were unfamiliar with the concept. To address this, we created a base set of features that demonstrated high importance in many use cases, and automatically include these features in all newly created experiments. Thus, additional features are recommended but not required for new use cases.  Furthermore, the study highlighted the need for high-quality documentation and tool-tips to guide the user through the platform.

Recently, a product team evaluated several different personalization methods: manual creation and hand-tuning of models; a manual conversion of experimental data to personalized methods toolbox; and the self-serve \PEX platform. The team evaluated the performance of these methods in the context of the necessary time investment and the improvement to product metrics of interest. \PEX out-performed the toolbox in both time investment and product impact. In comparison to hand-tuned models, \PEX performed on par to slightly worse with respect to metric impact, but is substantially easier to use; hand-tuning models is approximately 6 months per use case, while the active engineering effort for using \PEX is approximately 2-3 weeks per use case. Ultimately, due to the scale of use cases that can be explored via \PEX and the relatively good performance for product metrics, the team chose to heavily invest in \PEX for personalization needs.

\eat{
Usage of the \PEX platform for personalization of end-user product experience is standard in product growth teams at Facebook. Typically in these types of use cases, the team is interested in finding an optimal tradeoff between end user engagement and a counter metric such as SMS cost or capacity resources. Identifying optimal trade-off points manually can be quite time intensive as there are many permutations of treatment assignments, and as discussed above hand-tuning models is also time consuming. \PEX provides a simple, intuitive approach to ML based personalization via automatic data collection, monitoring and alerting, management of data pipelines and hosting, and automatic interpretable ML analysis.
}

We now review how a typical deployment experience for \PEX clients supports self-serve. Clients discover the \PEX platform, via pull or push marketing, and initialize their use case from the integrated UI flow that guides them through selection of treatment groups, product metrics \eat{(used as labels in our underlying models)}, and features. Most clients complete setup independently but some request an initial consultation with the platform team to discuss the fit of their use case. After 1-2 weeks, the product engineer returns to the integrated UI to review the offline counterfactual policy evaluation which provides product-metric impact analysis and automatic user segmentation analysis for enhanced interpretability. Again most users are able to move independently from offline testing to online testing, but those with weaker analytical background prefer to meet with the platform team to review the offline results. During product launch, many clients feel more reassured with a review from the platform team after they independently complete the launch flow and necessary code changes. The platform team provides support via weekly office hours and Q\&A fora, but also collaborates
with product teams on high-impact high-complexity applications and may implement additional platform functionalities on request.
\eat{
 This is mutually beneficial as the product team receives high levels of direct support, and the platform team is able to empirically evaluate new methods in production.
}

\begin{figure*}[h]
\includegraphics[width=0.7\paperwidth]{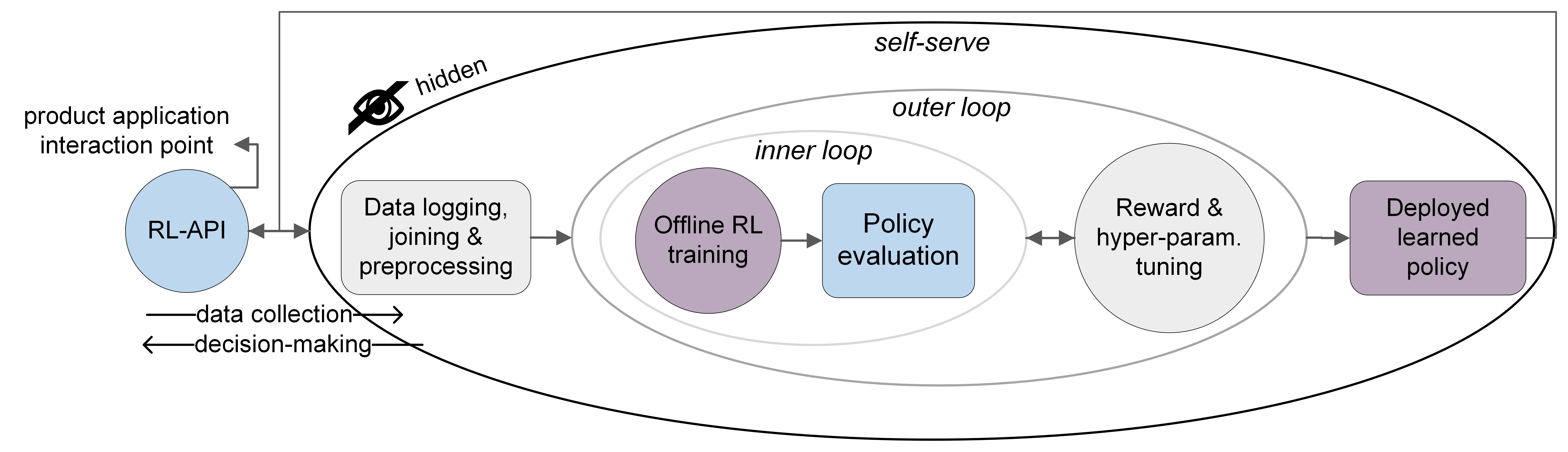}
\vspace{-5mm}
\caption{
Our workflow for self-serve reinforcement learning (RL).
\label{fig:RL}
}
\end{figure*}

\subsection{Self-serve reinforcement learning}
\label{sec:rl}

Reinforcement Learning (RL) is used in applications driven by metrics that are {\em delayed} from actions they evaluate and/or are {\em cumulative} over many actions (long-term). Using Supervised Learning in such cases is problematic because adequate labels are lacking.
Trained ML policies generally process outputs of ML models (trained on observational tasks) and are (manually) configured via dedicated parameters.
There is also significant risk of unintentional {\em negative feedback loops} undermining long-term product impact. In contrast to {\em ad hoc} policies for traditional supervised learning,
RL parameterizes policies explicitly and optimizes them \cite{sutton2018reinforcement}. RL can also track feedback loops and sequential dependencies while optimizing long-term product impact. Traditionally, the RL agent interacts with the environment over the course of learning ({\em online learning}). But in industry applications, this is often expensive, risky, etc.

\noindent
\textbf{{\em Offline reinforcement learning}} \cite{levine2020offline} reduces these risks and is practical in many applications. Using static logged datasets as in supervised learning,
offline RL extracts better policies via {\em off-policy learning} by generating trajectories that may have never been observed in the dataset. This makes offline RL relevant to industry applications where various interactions of previously deployed policies and systems have been logged.
But higher software and mathematical complexity keeps offline RL challenging for production applications. The rare success stories often include extensive hand-tuning and difficult maintenance.
%
Current efforts focus on training and keep in mind extrapolation errors seen in off-policy learning. Despite promising results in RL literature, deploying self-serve industry applications
on ML platforms would require additional AutoML components and must address several remaining challenges:%

\begin{itemize}
\item
{\bf Quality datasets and the need for randomization.} Without environmental interaction during learning, the logged dataset
must offer good quality and sufficient coverage. In healthcare and autonomous driving applications, it is difficult to ensure coverage,
but recommendation systems and Web-based services are less sensitive
to randomized exploration, which can be used to improve coverage. At the same time, logged datasets may be limited to some audiences or by manually drafted rules based on domain insights. Therefore, a self-serve offline RL system may need to augment prior historical data or collect fresh data.
\item
{\bf Effective training and evaluation pipelines.} Compared to supervised learning, workflows for RL in product applications are more difficult to configure and maintain. Self-serve offline RL requires effective and configurable pipelines for model training, evaluation and tuning to product metrics.
\item
{\bf Reward shaping.} When using RL, designing the reward function is commonly the first step. Offline RL for product applications is particularly sensitive to reward functions that must proxy for product metrics. Defining faithful proxies is challenging in practice since product applications often entail several product metrics with multiway different tradeoffs. Although metric values are normally found in the dataset logged to train offline RL, combining them in a reward function requires ($i$) application insights and ($ii$) mapping insights to a numerical reward function. This step is difficult to automate in self-serve platforms.
\end{itemize}

\begin{table*}
\caption{\label{tab:future} Future efforts outlined.}
\begin{tabular}{|c|c|}

\hline
\sc Future effort categories &
\sc Targets \\
\hline
\hline
\parbox{7cm}{
{\bf Proactive improv’t of platform adoption} (includes usability issues, but not ML technology - see below)
}
&
\parbox{9cm}{
Effective UIs, multiple APIs, back-end support, open architecture, documentation and tutorials, advocacy.
}
\\
\hline
\parbox{7cm}{
{\bf MLOps}
(stability, robustness, avoiding product metric regressions, moderate internal maintenance)
}&
\parbox{9cm}{
Reliability, resilience, effective maintenance, eng excellence and beyond (including ML eng excellence)
}\\
\hline
\parbox{7cm}{
{\bf Technology- and ML-driven development}
(high-value use cases,
new use scenarios for greater adoption, ML performance, effective resource usage)
}
&
\parbox{9cm}{
Support for large/high-value ML models. New individual components and services. Integration within the platform and with other platforms, product and platform metrics. Support for additional ML tasks, applications and use scenarios;  more effective ML models and decision policies; whole-strategy optimizations.
}
\\
\hline
\end{tabular}
\end{table*}

Our self-serve offline RL workflow Figure \ref{fig:RL} is implemented on both  of our platforms for clients without prior engineering expertise in RL. The level of automation we provide for RL is similar to that for supervised learning. A dozen of preexisting and new product applications at \Meta have onboarded to our RL workflow or started onboarding in the first month of availability. Application categories include Web-based services and
medium-scale recommendation systems.
Several product applications have already improved product metrics by $40-50$\% over prior production policies based on supervised learning models.

\noindent
\textbf{The self-serve client journey} includes: (1) data logging, (2) data preprocessing for RL, (3) policy training; evaluation by product metrics rather than RL-centric metrics,
(4) reward shaping and hyperparameter tuning (5) online deployment, (6) recurrent training.

\noindent
\textbf{To onboard} a product application, an owner specifies product metrics to be optimized and prediction features for the product application. These steps take under one hour using the integrated platform UI. The next and final step is for the application owner to extend the software-defined RL API for use in their product code. The product application uses this API for data collection, training and online deployment of any learned policy with offline RL. In detail, data collection starts with a randomized policy or some preexisting application-specific policy. Data are collected and preprocessed in the appropriate RL schema of tuples of one-step state transitions. After sufficient coverage of every action has been obtained in the collected data, offline RL training starts. Our system is integrated with ReAgent, an open-source RL platform, that offers well tested state-of-the-art RL algorithms. Trained policies are evaluated using {\em counterfactual analysis} which estimates impact on product metrics if the policy be deployed online. Training and evaluation are driven by the {\em reward function} specified by the application owner as a proxy of product metrics. Commonly, no such function is known up-front, and thus our {\em reward-tuning workflow} constructs one by exploring the space of linear parameterized reward functions by means of Bayesian Multi-Objective Optimization. Linear combinations of products metrics with different weights of preference are compared to identify non-dominated (Pareto optimal) learned policies. The same methodologies of Bayesian Multi-Objective Optimization and counterfactual evaluation are leveraged by our system for every learning-based tuning, e.g., model architectures, hyperparameters, etc. Candidates produced by reward tuning/training are reviewed by the owner, and based on how the different learned policies impact the product-metrics, the most promising one is selected for online deployment. After that, our system continues to collect training data based on the deployed policy with a small allowance for exploration (vs. exploitation), optionally controllable by the owner based on the application requirements. Collected data enables recurrent training.

\noindent
\textbf{From a self-serve perspective}, our offline RL system is application-agnostic but its owner-side invocation is RL-agnostic and requires no RL expertise.
Automation and tuning are enabled for end-product metrics rather than in terms of RL's internal metrics or proxy objectives, as is common in the literature \cite{paine2020hyperparameter, kumar2021workflow}. Owners of product applications can set up their use case and deploy an offline RL learning policy online in a few days, assuming sufficient application traffic.
The value of our self-serve RL workflow is in automation, as other RL solutions 
often require one-off coding and tuning that are hard to generalize and maintain in the long term.

\section{Conclusions and Future Work}
\label{sec:conclusions}

Our long-term vision for building self-serve ML platforms is developed by examining client journeys, identifying tasks that require unnecessary manual work, and automating this work by improvements in UI or system back-end, or by implementing relevant AutoML techniques. Our experience with end-to-end ML platforms suggests that the notion of self-serve that we develop helps lower the barrier for adoption of ML in product applications, broaden this adoption, reduce the time to product impact, and reduce maintenance costs. Product engineers with limited ML experience directly benefit from self-serve platforms, but questions remain about power-users who understand the underlying techniques and are often interested in finer control over individual tasks. This is why we advocate open-architecture end-to-end ML platforms to optionally automate selected parts of the overall workflow, e.g., any of data ingestion, training, serving, monitoring, etc. A power user interested in tweaking a particular task for high-value ML models and applications would rely on the automation and reproducibility of other tasks. When improving the performance of ML models, there is a risk of focusing too much on the model’s loss function and losing sight of product impact and overlooking powerful optimizations beyond the loss function of an ML model. An end-to-end platform can guide clients appropriately and make necessary tools available to them. Moreover, in some applications, it is more important to build and deploy additional use cases than super-optimize existing use cases.

With cliemts in mind, our work explains how to improve end-to-end ML platforms to help optimize key product metrics and ensure broader adoption by attaining the self-serve level. We foresee three categories of future efforts, as shown in Table \ref{tab:future}.
Each of these effort categories leverages AutoML in essential ways. Platform adoption can be proactively improved by automating typical points of failure for new clients and adding support for prototyping to decrease the cognitive effort needed to demonstrate the first success. Formalizing and automating MLOps workflows helps improve platform reliability and reduce maintenance costs for clients and platform owners. AutoML is also essential to enable more sophisticated ML models and reduce the configuration burden for clients of such models.

Self-serve platforms amplify the impact of end-to-end platforms and help accumulate “passive income” from use cases deployed on the platform over long periods of time. However, the broad impact of a given development effort can be difficult to estimate (beyond pilot clients) because it depends on preferences of many clients and because several efforts often contribute to any one result. We distinguish the needs of specific clients from platform improvements initiated by the platform team with numerous clients in mind. Both types have been helpful in the past and will be pursued in the future. Another strategic dichotomy pits a handful of sophisticated high-value models against numerous simpler “tail models” that support various aspects of non-ML-driven applications. While open architectures and partial workflows should be helpful for power users, it makes sense to (i) consider client tiers by support needed, (ii) prioritize clients by their likely impact on product applications.

\bibliographystyle{ACM-Reference-Format}
\bibliography{self-serve}

\end{document}